\documentclass[10pt,letterpaper]{article}

\usepackage{cogsci}

\cogscifinalcopy 

\usepackage{amsmath}
\usepackage{pslatex}
\usepackage{apacite}
\usepackage{float} 
\usepackage{linguex}
\usepackage{graphicx}
\usepackage{url}
\usepackage{CJKutf8}

\newcommand{\key}[1]{\textsc{#1}}

\title{Availability-Based Production Predicts Speakers' Real-time Choices of Mandarin Classifiers}
 
\author{{\large \bf Meilin Zhan (meilinz@mit.edu)} \ \ {\large \bf Roger Levy (rplevy@mit.edu)} \\
  Department of Brain \& Cognitive Sciences, Massachusetts Institute of Technology \\
  43 Vassar Street, Cambridge, MA 02139 USA}

\begin{document}

\maketitle

\begin{abstract}
  Speakers often face choices as to how to structure their intended message into an utterance.  Here we investigate the influence of contextual predictability on the encoding of linguistic content manifested by speaker choice in a classifier language, Mandarin Chinese. In Mandarin, modifying a noun with a numeral obligatorily requires the use of a classifier. While different nouns are compatible with different \key{specific} classifiers, there is a \key{general} classifier that can be used with most nouns. When the upcoming noun is less predictable, using a more specific classifier would reduce the noun's surprisal, potentially facilitating comprehension (predicted to be preferred under Uniform Information Density, Levy \& Jaeger, 2007), but the specific classifier may be dispreferred from a production standpoint if the general classifier is more easily available (predicted by Availability-Based Production; Bock, 1987; Ferreira \& Dell, 2000). Here we report a picture-naming experiment confirming two distinctive predictions made by Availability-Based Production.

\end{abstract}

\section{Introduction}

The simple act of speaking may typically seem effortless, but it is extraordinarily complex.  Speakers must plan the message they wish to convey, choose words and constructions that accurately encode that message, organize those words and constructions into linearly-sequenced utterances, keep track of what has been said, and execute each part of their speaking plans at the correct time. Throughout this process, speakers face choices in structuring their intended message into an utterance. One central question for a computationally precise theory of language production is thus: When multiple options are available to express more or less the same meaning, what general principles govern a speaker's choice? To what extent do speakers make choices that potentially facilitate comprehenders, and to what extent do they make choices that are preferable from a production standpoint? Here we approach these questions from the standpoint of contextual predictability, which is known to affect a wide range of speaker choices. Specifically, we investigate the influence of contextual predictability on the encoding of linguistic content manifested by speaker choice in a classifier language.  Two major theories of sentence production, Availability-Based Production (ABP; \citeNP{bock:1987, ferreira-dell:2000}) and Uniform Information Density (UID; \citeNP{levy-jaeger:2007, jaeger:2010}), make conflicting predictions about the distribution of speaker choices when more than one classifier could be used in a given context.  We report a language production experiment on classifier choice that adjudicates between these theories.

\begin{CJK*}{UTF8}{gbsn}

In languages with a grammaticalized count--mass distinction, such as English, count nouns such as \textit{table} can be used with a numeral directly and typically exhibit a singular--plural morphological marking (e.g., \textit{one table, three tables}), whereas mass nouns such as \textit{sand} cannot co-occur with numerals directly without some kind of measure word (e.g., \textit{three cups of sand}) and do not have a plural morphology on the noun (e.g., \textit{*three sands}).
%
%
In classifier languages such Mandarin, in contrast, nouns lack obligatory singular--plural morphological marking and cannot directly co-occur with numerals. Instead, a numeral classifier is required when a noun is modified by a numeral or a demonstrative. Linguists generally agree that there is a distinction between two types of Chinese classifiers: count classifiers, which we focus on here, and mass classifiers \cite{tai:1994, cheng-sybesma:1998, li-etal:2008}. \footnote{A count classifier (e.g., two CL.top hat (``two hats")) is used to categorize a class of noun entities in reference to their salient perceptual properties, which are often permanently associated with the entities named by the class of nouns. A mass classifier (e.g., two box (of) hat (``two boxes of hats")) creates a unit and form a temporary relationship with the noun. Because using different mass classifiers often change the semantics of the noun phrase, here we only focus on count classifiers (henceforth, classifiers).} Among count classifiers, which are used with nouns that denote individuals or groups of individuals, different \key{specific} classifiers are compatible with different nouns, but the \key{general} classifier \textit{ge} (个) can be used with almost any noun. Often, the choice of general versus specific classifier for a given noun carries little to no meaning distinction for the utterance, as illustrated in \ref{ex:tai} and \ref{ex:ge} below.

\exg. 我 卖了    三    \textbf{台}         电脑\\
      wo mai-le san   \textbf{tai}     diannao\\
      \small{I  sold three CL.machinery   computer (``I sold three computers")}\\ \label{ex:tai}

\exg. 我 卖了 三    \textbf{个}       电脑\\
      wo mai-le san   \textbf{ge}     diannao\\
      \small {I sold three CL.general  computer (``I sold three computers")}\\ \label{ex:ge}

\end{CJK*}

In this study, we focus on speaker choice between general and specific count classifiers for nouns where both options convey more or less the same meaning. When the upcoming noun is unpredictable, a specific classifier would constrain the range of possible nouns more than the general classifier, thus increasing the predictability of the upcoming noun and potentially benefiting comprehension.  The Uniform Information Density account thus predicts that speakers will prefer specific classifiers for unpredictable nouns.  However, Availability-Based Production predicts that the specific classifier may be dispreferred from a production standpoint if the general classifier is more easily available. Which of these two accounts better predict classifier choice in real-time production? In other words, does noun predictability affect classifier choice, and if so, in which direction? Here we use a picture-naming experiment to address this question. 

Before diving into the experiment, we first briefly introduce why we focus on predictability effects and how the two accounts predict speaker choices with regard to optional reduction in language. 

\section{Predictability Effects on Optional Reduction}

It has been shown that contextual predictability plays a role in optional reduction in language, where more predictable content tend to yield a greater rate of reduction in the linguistic form. At the lexical level, predictable words are phonetically reduced \cite{jurafsky-etal:2001, bell-etal:2009, seyfarth:2014} and tend to have shorter forms \cite{piantadosi-etal:2011, mahowald-etal:2013}. At the syntactic level, optional function words are more likely to be omitted when the phrase they introduce is predictable \cite{levy-jaeger:2007, jaeger:2010}. For example, in English relative clauses (henceforth RCs) such as \ref{ex:rc1}, speakers can but do not have to produce the relativizer \textit{that}. We refer to the omission of \textit{that} as \textsc{optional reduction}.

\ex. I created a mobile app dancers like. \label{ex:rc1}

\ex. I created a mobile app that dancers like. \label{ex:rc2}

For optional function word omission, predictability effects have been argued to be consistent with both the speaker-oriented account of Availability-Based Production, where the speaker mentions material that is readily available first, and the potentially audience-oriented account of Uniform Information Density, where the speaker aims to convey information at a relatively constant rate. These two accounts have proven difficult to disentangle empirically. For different reasons, both accounts predict that the less predictable the clause introduced by the function word, the more likely the speaker would be to produce the function word \textit{that}.

\subsection{Uniform Information Density}

Uniform Information Density proposes that within boundaries defined by grammar, when multiple options are available to express the message, speakers prefer the variant that distributes information density more uniformly throughout the utterance, to lower the chance of information loss or miscommunication \cite{levy-jaeger:2007, jaeger:2010}. Multiple formalizations are possible under this account \cite{genzel-charniak:2002,aylett-turk:2004,maurits-etal:2010,levy:2018cogsci}.

In \ref{ex:rc1}, where the relativizer \textit{that} is omitted, the first word of the relative clause $w_1$ (\textit{dancers} in this case) is highly unpredictable and would convey two pieces of information: both the onset of the relative clause and part of the content of the relative clause itself. These both contribute to the information content of $w_1$, which can be measured using \textsc{surprisal}, the negative log-probability of the word in context: $- \log P(w|\text{Context})$ \cite{hale:2001, levy:2008, demberg-keller:2008, smith-levy:2013}. In \ref{ex:rc2}, having \textit{that} at the onset of the RC splits these two pieces of information apart, offloading the relative clause's onset onto \textit{that} so that \textit{dancers} only conveys relative clause-internal content and thus has lower information content, potentially avoiding a peak in information density and thus facilitating comprehension.

\subsection{Availability-Based Production}

Availability-Based Production proposes that production is more efficient if speaker mentions material that is readily available first. According to ABP, speaker choice is governed by: 1) when a part of a message needs to be expressed within an utterance; 2) when the linguistic material to encode that part of the message becomes available \cite{bock:1987, ferreira-dell:2000}. Specifically, if material that encodes a part of the message becomes available when it comes time to convey that part of the message, it will be used. However, if that material is not yet available, then other available material will be used, as long as it is compatible with the grammatical context produced thus far and it does not cut off the speaker's future path to expressing the desired content. This is also referred to as \key{the principle of immediate mention} \cite{ferreira-dell:2000}.

Suppose a speaker has just uttered the word \textit{app} in \ref{ex:rc1} and has in mind to convey the remainder of the utterance meaning as a relative clause. If the word \textit{dancers} becomes available quickly, then according to the principle of immediate mention, a sentence without \textit{that} should be produced (see \ref{ex:rc1}).  If \textit{dancers} does not become available quickly, however, ABP predicts that the speaker will utter \textit{that} to buy more time for \textit{dancers} to become available.  (Note that this account relies on an implicit auxiliary assumption that that \textit{that} will generally become available quite quickly; this assumption is rendered plausible by the fact that it is a high-frequency word used in a wide variety of contexts.)  If the first word of the RC takes longer to become available the lower its contextual predictability---an assumption consistent with previous work on picture naming \cite{oldfield-wingfield:1965} and word naming \cite{balota-chumbley:1985}---then the less predictable the relative clause, the lower the probability that its first word, \textit{dancers}, will be available at  when the speaker reaches the RC, and the higher the probability that the speaker will use \textit{that}.  Since an RC is required after \textit{app} in order for it to be followed by the word \textit{dancers}, the lower the contextual probability of an RC the lower the contextual probability of its first word, predicting the empirically observed relationship between phrasal onset probability and optional function word omission rate.

\subsection{Distinguishing theories of predictability-driven speaker choice}

Although UID and ABP are substantially different theories of what drives speaker choice, they make the same prediction for the effect of contextual predictability on optional reduction of function words for cases such as \ref{ex:rc1}. It is thus intrinsically difficult to use optional reduction phenomena to tease these accounts apart. Prior work \cite{jaeger:2010} acknowledged this entanglement of the predictions and attempted to tease these accounts apart via joint modeling using logistic regression. There are other phenomena for which the accounts make similar predictions, as well. Consider the case of ordering choices for words or phrases, such as subject--object versus object--subject word order for languages in which both options are available, such as Russian.  Availability-Based Production predicts that whichever becomes available earlier will be uttered first \cite{levelt-maasen:1981-lexical-search}; if the lexical encodings of more contextually predictable references tend to become available more quickly, then more predictable arguments will tend to be uttered first.  This prediction is indeed likely to be true: a given-before-new word order preference is widely recognized to influence many languages \cite{behaghel:1930-deutscher-wortstellung,prince:1981towards,gundel:1988universals}, and discourse-given entities are generally more contextually predictable than discourse-new entities.  But UID turns out to make the same prediction.  Two arguments of the same verb generally carry mutual information about each other, so any argument will typically be less surprising if it is the latter of the two.  Thus, putting the argument that is more predictable from sentence-external context before the less-predictable argument will lead to a more uniform information density profile and will be preferred.

In the case of speaker choice for Mandarin classifiers, however, UID and ABP turn out to make different predictions as we describe in the next section.  The empirical facts regarding speaker choice for classifiers are thus of considerable theoretical interest.

\begin{CJK*}{UTF8}{gbsn}

\section{Predictions on Mandarin Classifiers}

%
%
%

\end{CJK*}

\citeA{zhan-levy:2018naacl} have argued that UID and ABP make different predictions on Mandarin Classifier use with regard to noun predictability. As regards UID, the choice between a specific classifier and a general classifier will typically affect the contextual predictability of the noun modified by the classifier. In particular, a specific classifier constrains the space of possible upcoming nouns more tightly than the general classifier \cite{klein-etal:2012}, thus generally reducing the actual noun's surprisal. The UID hypothesis thus predicts that speakers choose a \textbf{specific} classifier more often when the noun predictability would otherwise be low than when the noun is more predictable. This is because the use of a specific classifier makes the distribution of information density more even between the noun and the classifier.


Availability-Based Production, on the other hand, makes different predictions than UID. The fundamental prediction of ABP is that the harder the noun lemma is to access, the less often the speaker will use a specific classifier, provided two plausible assumptions. First, the general classifier \textit{ge} is always available, regardless of the identity of the upcoming noun, as it is the most commonly used classifier and is compatible with practically every noun.  Second, in order to access and produce an appropriate specific classifier, a speaker must complete at least some part of the planning process for the production of the nominal reference: accessing the noun lemma, or minimally accessing the key semantic properties of the referent that determine its match with the specific classifier. On these two assumptions, any feature of the language production context that makes the noun lemma less accessible or that more generally makes noun planning more difficult will favor the general classifier.  In out-of-linguistic-context picture naming, for example, noun lemma accessibility is known to be driven by noun frequency \cite{oldfield-wingfield:1965}. The lower the noun frequency, the less accessible the noun lemma, thus the less likely a specific classifier will be used.  To make predictions about the effect of noun predictability on classifier choice in linguistic contexts, we must add a third, theoretically plausible assumption: that less predictable noun lemmas are harder and/or slower to access than more predictable noun lemmas. 
On these three assumptions, in corpus data the link between noun lemma accessibility and classifier choice will show up as an effect of noun predictability, which by hypothesis is determining noun lemma accessibility. For less predictable nouns, their specific classifiers will less likely be available to the speaker when the time comes to initiate classifier production. Because noun lemmas need to be accessed in order to produce specific classifiers, and the less predictable the noun, the harder the noun lemma is to access and hence the specific classifier associated with the noun becomes available by the time a classifier needs to be produced.

In other words, the link between noun lemma accessibility and classifier choice will manifest in different predictions depending on whether one we are looking at usage in linguistic context versus picture-naming.  Under our assumptions about ABP, we can identify three predictions.  First, in out-of-context picture naming, speakers should as described above choose the \textbf{general} classifier more often the more frequent the noun (provided there is high naming agreement for the picture, so that there is not competition among nouns that affects the production process).  Second, in corpus data, speakers should as described above choose a \textbf{general} classifier more often the less predictable the noun. Finally, we can add a third prediction based on the temporal dependence of specific classifier production on noun planning: when speakers are under greater time pressure, they should produce the general classifier more often, as it can be used even when noun planning has not proceeded far enough for a specific classifier to be available.

\citeA{zhan-levy:2018naacl} tested the second prediction in an investigation of naturally occurring texts, using language models to estimate noun predictability and mixed logistic regression to infer its relationship with classifier choice.  They found that the less predictable the noun, the lower the rate of using a specific classifier. While these results lend support for the Availability-Based Production account, the study has some limitations. One limitation is that the corpus being used was a collection of online news texts. Written language may serve as a first approximation of testing theories of language production, but it would be ideal to use real-time language production task to further test the hypotheses. Another limitation is that there was no experimental control of context, so it is possible that  predictability was confounded with some other contextual factor that was not included in their regression analysis but that is actually responsible for speaker choice. 

In the present study, we use a real-time language production task involving picture naming varying noun frequency and whether speakers are put under time pressure, allowing us to further investigate the two models of language production by testing the first and third predictions described above.

\section{Methods}
We used a picture-naming experiment to test the predictions of Uniform Information Density and Availability-Based Production by manipulating noun frequency and whether or not the speaker is under time pressure. This picture-naming experiment offers a simple yet effective way to elicit real-time language production.

\subsection{Participants} Thirty-six self-reported native speakers of Mandarin Chinese were recruited via Witmart, a China-based online crowdsourcing platform. Participants received compensation for their time.

\subsection{Materials}
We adapted images from the Pool of Pairs of Related Objects (POPORO) \cite{kovalenko-etal:2012} image set to create our visual stimuli. We selected images from the image set based on the following criteria: 1) the image can be described by a count noun; 2) the preferred count noun is compatible with the general classifier and at least one specific classifier. We developed a web-based version of the experiment using \texttt{jsPsych} \cite{deleeuw:2015}, a JavaScript library for creating behavioral experiments in a web browser.  We estimated the frequencies of occurrence of the preferred count nouns from  \texttt{SogouW} \cite{sogou:2006}, a word frequency dictionary for online texts in Chinese.

\begin{figure}[t]
\begin{center}
\includegraphics[scale=0.62]{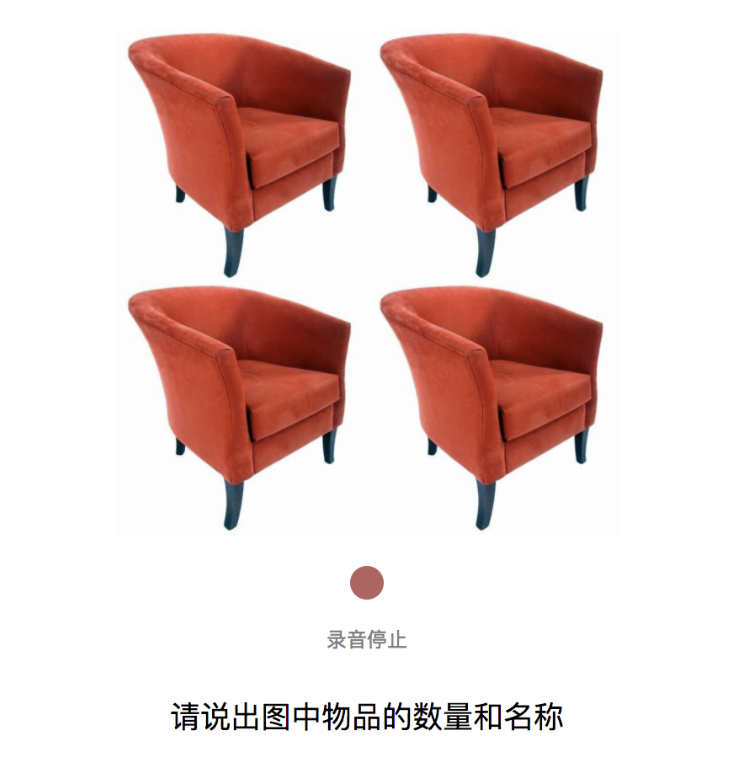}
\end{center}
\caption{Sample visual display for the picture-naming experiment. The red dot below the picture is the recording light. Below it is the text indicating the status of the recorder, in this case, it is "recording stopped". The English translation for the sentence in the bottom is: ``Please describe the number and the name of the objects in the picture." } 
\label{sample}
\end{figure}

\subsection{Procedure}
Participants were presented with scenes of various countable object kinds such as cabbages and tables. Figure \ref{sample} shows a sample display. In each scene, there were several instances of the same object kind. The number of objects in each trial varied from two to four. Participants were asked to describe the number and the name of the object in Mandarin, eliciting utterances such as ``three CL chairs" which we recorded. 

Participants were assigned to one of the two conditions. In the \textbf{Quick} condition, recording started 50 ms after the picture was shown, indicated by a recording light at the bottom of the picture. Each trial ended after 5 seconds of recording, and the next trial began automatically. In the \textbf{Slow} condition, recording stared 3 seconds after the picture was shown, and participants clicked on the screen to move toward the next trial. 

\subsection{Predictions}

Availability-Based Production predicts that (1) the rate of specific classifier use will be lower in the \textbf{Quick} condition, when speakers are under time pressure, than in the \textbf{Slow} condition; and (2) the rate of specific classifier use will be lower for less frequent nouns.  This latter prediction derives from evidence that lexical access, as manifested by response latencies, takes longer for lower frequency words in language production experiments requiring word production outside of sentence context; this holds not only for picture naming \cite{oldfield-wingfield:1965}, as we require of participants here, but also of visually-presented word naming \cite{balota-chumbley:1985}.  If lower-frequency nouns are slower to access, their specific classifiers may also be slower to access and thus less often used than the general classifier, which is available for all nouns.

The predictions of Uniform Information Density for the effect of the \textbf{Quick}/\textbf{Slow} manipulation are unclear.  As regards noun frequency, UID predicts that if anything low-frequency nouns should have a \emph{higher} rate of specific classifier usage, as a noun's frequency may effectively serve as its predictability in this experimental setting without broader linguistic context.

\begin{figure}[t]
\begin{center}
\includegraphics[scale=0.62]{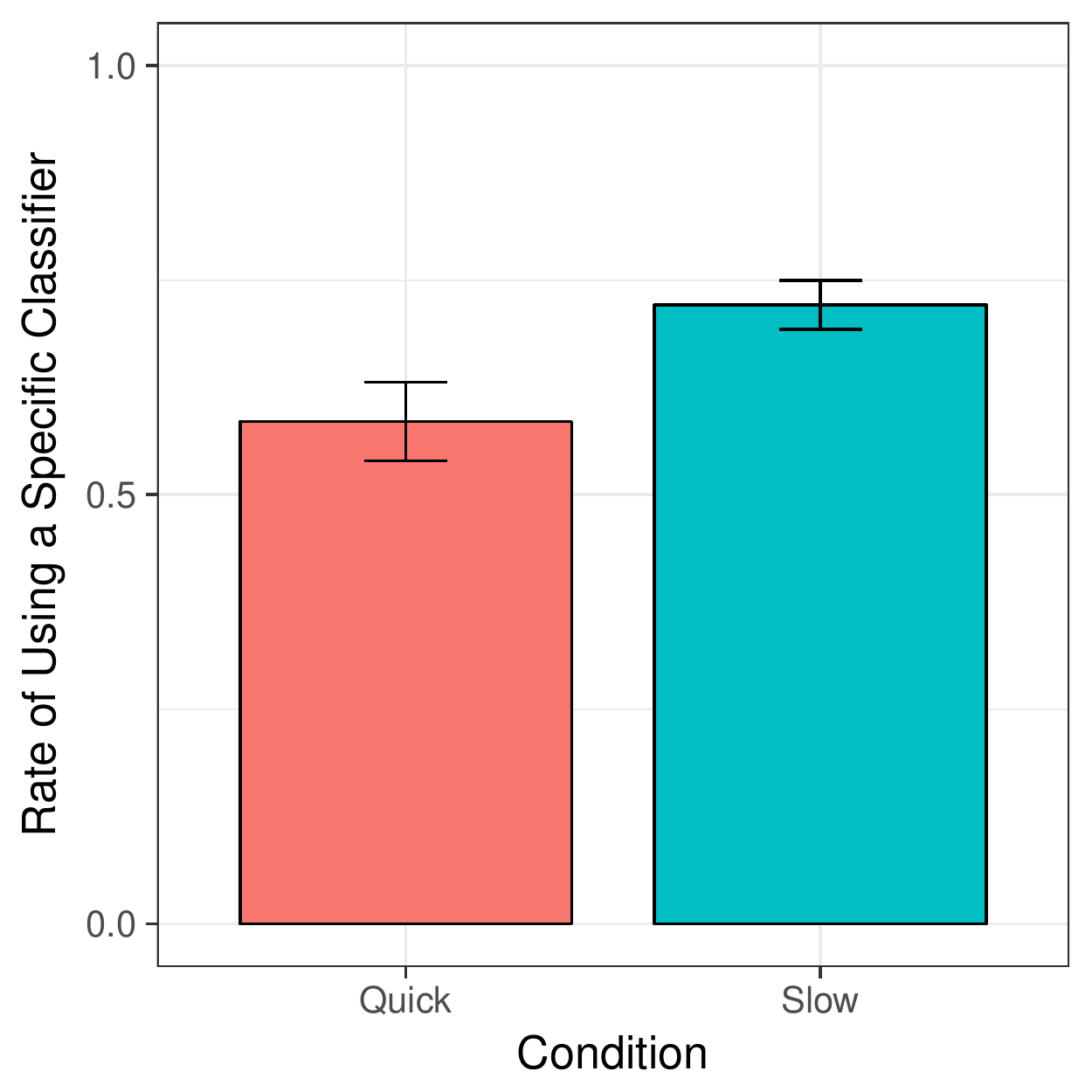}
\end{center}
\caption{Quick vs. Slow manipulation and rate of using a specific classifier as opposed to the general classifier \textit{ge}. Error bars are standard errors over by-participant means.} 
\label{result_bar}
\end{figure}

\begin{figure}[t]
\begin{center}
\includegraphics[scale=0.62]{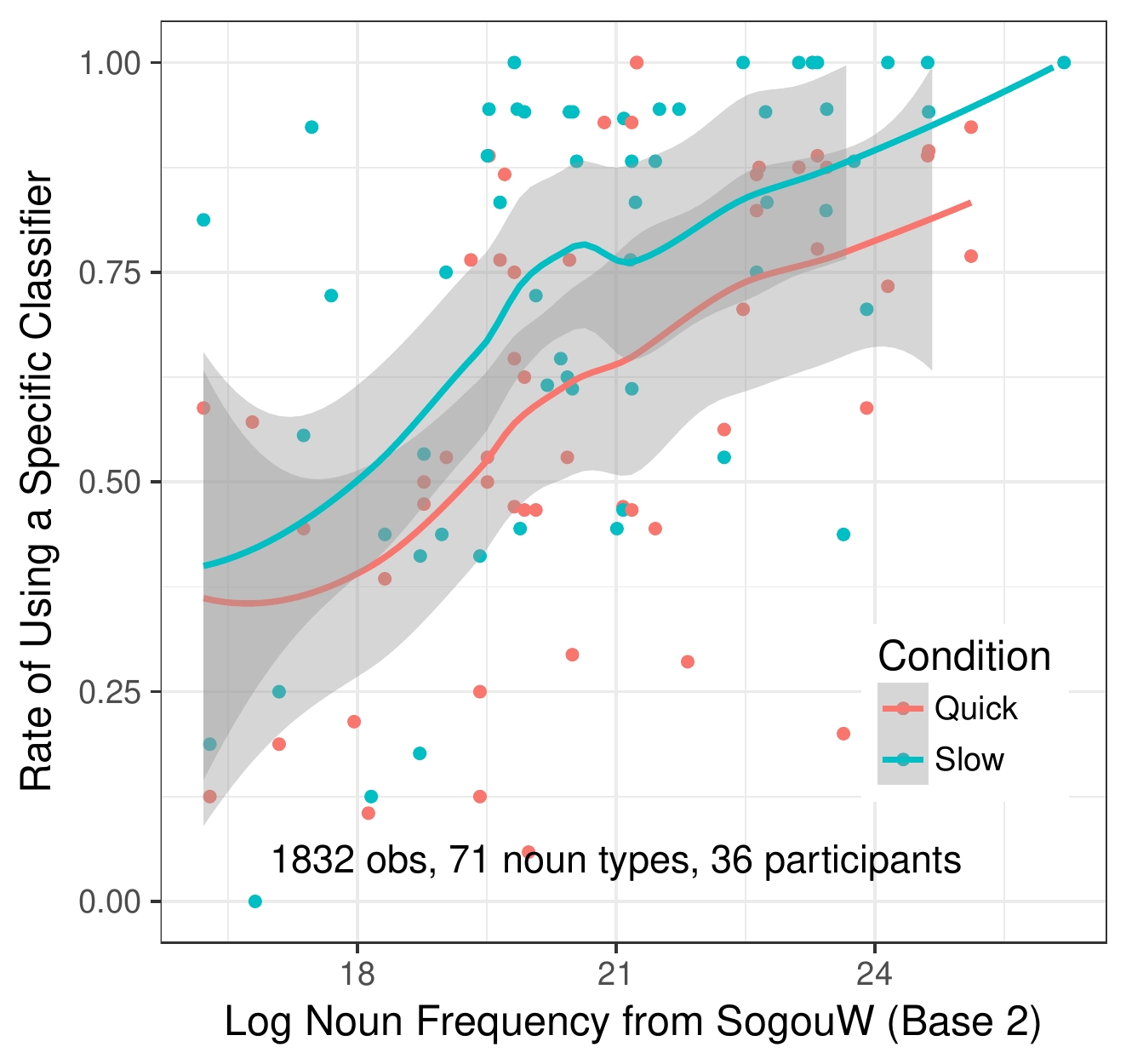}
\end{center}
\caption{The relationship between noun frequency and rate of specific vs.\ general classifier use in picture naming.} 
\label{result}
\end{figure}

\subsection{Analysis}
Audio responses were first transcribed to texts using Google's speech-to-text application programming interface (API), and then checked manually to correct transcription errors. We excluded trials when the participant did not produce a classifier or a noun. For each item, we used the nouns that were most frequently produced as the noun for that item. We also compiled a list of acceptable nouns for each items, and excluded nouns that were not on the list.

We used a mixed-effect logit model to investigate whether noun frequency and time pressure affect classifier choice. The dependent variable was the binary outcome of whether the general classifier or a specific classifier was produced. For each noun type, we also identified its preferred specific classifier (using native speaker introspective judgment and predominant responses by experimental participant, which were concordant). We included two predictors in the analysis: log noun frequency and condition. We included noun, preferred specific classifier, and participant as random factors. We used the maximal random-effects structure with respect to these two predictors \cite{barr-etal:2013}.  For condition, this entailed random slopes by noun and by preferred specific classifier, but not participant because the condition manipulation was between subject. For log noun frequency, this entailed a random slope by participant. The full formula in the style of \texttt{R}'s \texttt{lme4} is:

\begin{quote}\begin{quote}
{\small\verb!response ~ log_noun_freq + condition!} {\small\verb!+(1+condition|noun)!} 
{\small\verb!+(1+condition|preferred_spec_cl)!} {\small\verb!+(1+log_noun_freq|participant)!}
\end{quote}\end{quote}

Statistical significance was determined using Markov chain Monte Carlo (MCMC) methods in the R package \texttt{MCMCglmm} \cite{hadfield:2010} with $p$-values based on the posterior distribution of regression model parameters with an uninformative prior, as is common for MCMC-based mixed model fitting \cite{baayen-etal:2008}. 

\section{Results}

Looking just at the \textbf{Quick}/\textbf{Slow} contrast, we find (Figure \ref{result_bar}) that speakers produced more instances of the general classifier when they are under time pressure than when they are not ($p < 0.05$), suggesting that specific classifiers are slower than the general classifier to access and thus supporting the Availability-Based Production account.

Further breaking out our results by noun log-frequency, we find  (Figure \ref{result}) that the lower frequency the noun, the more likely a \textbf{general} classifier is to be produced ($p < 0.001$). This pattern holds within both experimental conditions and is consistent with previous results from the corpus analysis \cite{zhan-levy:2018naacl}, and also supports the Availability-Based Production account. 

One potential concern arises in the frequencies of the different specific classifiers. One could argue that it was not the noun's frequency that determined the use of the general classifier, rather it was the frequency of the preferred specific classifier that affected the choice of which classifier was used. In the mixed-effect logit model presented above, we included a by-specific-classifier random intercept, which largely rules out the possibility that specific classifier frequency were confounding the effect of noun frequency. To further investigate this issue, we tried a version of our regression model that also includes a fixed effect for the log frequency of preferred specific classifier as a control factor. We did not find any qualitative change to the results. The effects of noun frequency ($p < 0.001$) and condition ($p < 0.05$) on classifier choice remain qualitatively similar to the results of the original model. Furthermore, in this new analysis, there is no effect of specific classifier frequency on classifier choice ($p = 0.483$). This additional analysis suggests that it is unlikely that specific classifier frequency to be driving the effect of noun frequency. 

\section{Conclusion}
Using a picture-naming experiment, we show that Availability-Based Production predicts speakers' real-time choices of Mandarin Chinese. The lower a noun's frequency, the more likely a general classifier is to be used. We also found that the use of classifier is moderated by whether the speaker is under time pressure when speaking, where the speaker tends to produce more instances of the general classifier if they are under greater time pressure to speak. This real-time effect confirms that the general classifier is easily accessible when the speaker is about to produce a noun phrase with numeral. 

Taken together, the present study and previous corpus work on Mandarin classifier \cite{zhan-levy:2018naacl} offer converging evidence regarding the relationship between noun frequency, predictability, and classifier choice, and thus shed light on the mechanisms influencing speaker choice. While the corpus work provides ecological validity through naturalistic data, the experimental work helps us to eliminate potential correlation-based confounds with a clean setup, and enables us to get dense data that are theoretically important but naturalistically sparse. When combined together, this work is complementary with previous corpus work and together paint a more comprehensive picture of language production. 

These studies also underscore the importance of investigating a wide variety of speaker choice phenomena, taking advantage of the many types of  phenomena offered by the languages of the world.  Optional reduction and word order choice are perhaps the best-studied types of such alternations, but they have proven ill-suited to teasing apart the predictions of Uniform Information Density and Availability-Based Production.  The approach taken here could be extended to the many types of classifier systems in languages around the world, and might inspire investigation of yet different speaker choice configurations that shed new insights into the mechanisms of language production.

In future work on classifier choice, we plan to investigate other potentially relevant factors such as mutual information. It is possible that some classifier-noun pairs are especially prominent and accessible in memory. If the mutual information between the noun and classifier is high, speakers might be more likely to use that classifier for the noun selected. Although we have not found direct evidence supporting the UID hypothesis, it is possible that this particular experimental setting is not very communicative in nature. In future work, we plan to do an real-time language production experiment in a more communicative setting, with virtual or real listeners in the experiment to further test speaker choice in language production. We also plan to add additional production measures such as phonetic reduction of classifiers, pause durations, and disfluencies to enrich our understanding of language production. 

Viewed most broadly, using speaker choice in classifier production as a test case has helped us investigate computationally explicit theories of language production, and advance our understanding of the psychological processes involved in converting our thoughts to speech.

\section{Acknowledgements}
We gratefully acknowledge valuable feedback from members of MIT Computational Psycholinguistics Laboratory and three anonymous reviewers, technical advice on web-based experiment development from Jon Gauthier and Wenzhe Qiu. This research was funded by NSF grants BCS-1551866 to Roger Levy, and BCS-1844723 to Roger Levy and Meilin Zhan.

\bibliographystyle{apacite}

\def\thebibliography#1{\section*{References}
  \small
   \list
   {[\arabic{enumi}]}{\leftmargin \parindent
     \itemindent -\parindent
     \itemsep 0ex plus 1pt
     \parsep 0.1ex plus 1pt minus 1pt
     \usecounter{enumi}}
     \def\newblock{\hskip .11em plus .33em minus .07em}
     \sloppy\clubpenalty4000\widowpenalty4000
     \sfcode`\.=1000\relax}

\setlength{\bibleftmargin}{.125in}
\setlength{\bibindent}{-\bibleftmargin}

\bibliography{zhan_levy_cogsci2019.bib}

\end{document}